%% file: main.tex
\documentclass{Interspeech}

\interspeechcameraready

\title{Synthetic Data Generation for Phrase Break Prediction\\with Large Language Model}
\author[affiliation={1}]{Hoyeon}{Lee}
\author[affiliation={2}]{Sejung}{Son}
\author[affiliation={3}]{Ye-Eun}{Kang}
\author[affiliation={1}]{Jong-Hwan}{Kim}

\affiliation{}{NAVER Cloud}{South Korea}
\affiliation{}{NHN}{South Korea}
\affiliation{}{Yale University}{USA}
\email{yeon.lee@navercorp.com}


\usepackage{comment}
\usepackage{multirow}
\usepackage{cite}
\usepackage{xcolor}
\definecolor{ForestGreen}{RGB}{0,170,0}
\definecolor{DarkGreen}{RGB}{0, 100, 0} 

\begin{document}

\maketitle
\input{contents/0-abstract}
\input{contents/1-introduction}

\input{contents/2-related}

\input{contents/3-method}
\input{contents/4-zerofew}

\input{contents/5-crosslingual}

\input{contents/6-xlb}

\input{contents/7-conclusion}

% \input{contents/8-ack}
% Acknowledgement should only be included in the camera-ready version, not in the version submitted for review. The 5th page is reserved exclusively for acknowledgements and  references. No other content must appear on the 5th page. Appendices, if any, must be within the first 4 pages. The acknowledgments and references may start on an earlier page, if there is space.

% \ifinterspeechfinal
%      The Interspeech 2025 organisers
% \else
%      The authors
% \fi
% would like to thank ISCA and the organising committees of past Interspeech conferences for their help and for kindly providing the previous version of this template.

\bibliographystyle{IEEEtran}
\bibliography{reference}

\end{document}

%% file: contents/0-abstract.tex


% Accurate phrase break prediction is crucial for enhancing the natural prosody of text-to-speech systems.
% However, current approaches heavily rely on substantial human annotations from audio or text, incurring significant manual effort and cost. 

\begin{abstract}
Current approaches to phrase break prediction address crucial prosodic aspects of text-to-speech systems but heavily rely on vast human annotations from audio or text, incurring significant manual effort and cost. Inherent variability in the speech domain, driven by phonetic factors, further complicates acquiring consistent, high-quality data. Recently, large language models (LLMs) have shown success in addressing data challenges in NLP by generating tailored synthetic data while reducing manual annotation needs. Motivated by this, we explore leveraging LLM to generate synthetic phrase break annotations, addressing the challenges of both manual annotation and speech-related tasks by comparing with traditional annotations and assessing effectiveness across multiple languages. Our findings suggest that LLM-based synthetic data generation effectively mitigates data challenges in phrase break prediction and highlights the potential of LLMs as a viable solution for the speech domain.
\end{abstract}

%Manuscripts submitted to Interspeech 2025 must use this document as both an instruction set and as a template. Do not use a past paper as a template. Always start from a fresh copy, and read it all before replacing the content with your own. The main changes with respect to previous years' instructions are \blue{highlighted in blue}.

% Before submitting, check that your manuscript conforms to this template. If it does not, it may be rejected. Do not be tempted to adjust the format! Instead, edit your content to fit the allowed space. The maximum number of manuscript pages is 5. The 5th page is reserved exclusively for acknowledgements and references, which may begin on an earlier page if there is space.

% The abstract is limited to 1000 characters. The one in your manuscript and the one entered in the submission form must be identical. Avoid non-ASCII characters, symbols, maths, italics, etc as they may not display correctly in the abstract book. Do not use citations in the abstract: the abstract booklet will not include a bibliography.  Index terms appear immediately below the abstract. 

%% file: contents/1-introduction.tex
\section{Introduction}
% 1. Background - (Multilingual) TTS 에서의 Phrase Break 중요성
Text-to-speech (TTS) is inherently a one-to-many mapping task, as one text input can yield multiple speech outputs depending on speaker and prosodic variations.
This ambiguity often leads to challenges in generating natural and contextually appropriate speech.
A key front-end module influencing this variability is phrase break prediction, which determines the appropriate pause placements between phrases.
Accurate phrase break prediction is essential for capturing the nuanced prosody that contributes to speech naturalness, especially in multilingual TTS where syntactic and prosodic rules vary significantly across languages~\cite{xiao2020improving,Futamata2021,Lee2023}.
To this end, previous studies have employed human annotations as a vital resource for defining phrase boundaries and capturing language-specific prosodic variations.

\begin{figure}[t]
    \centering
    \includegraphics[width=0.9\linewidth]{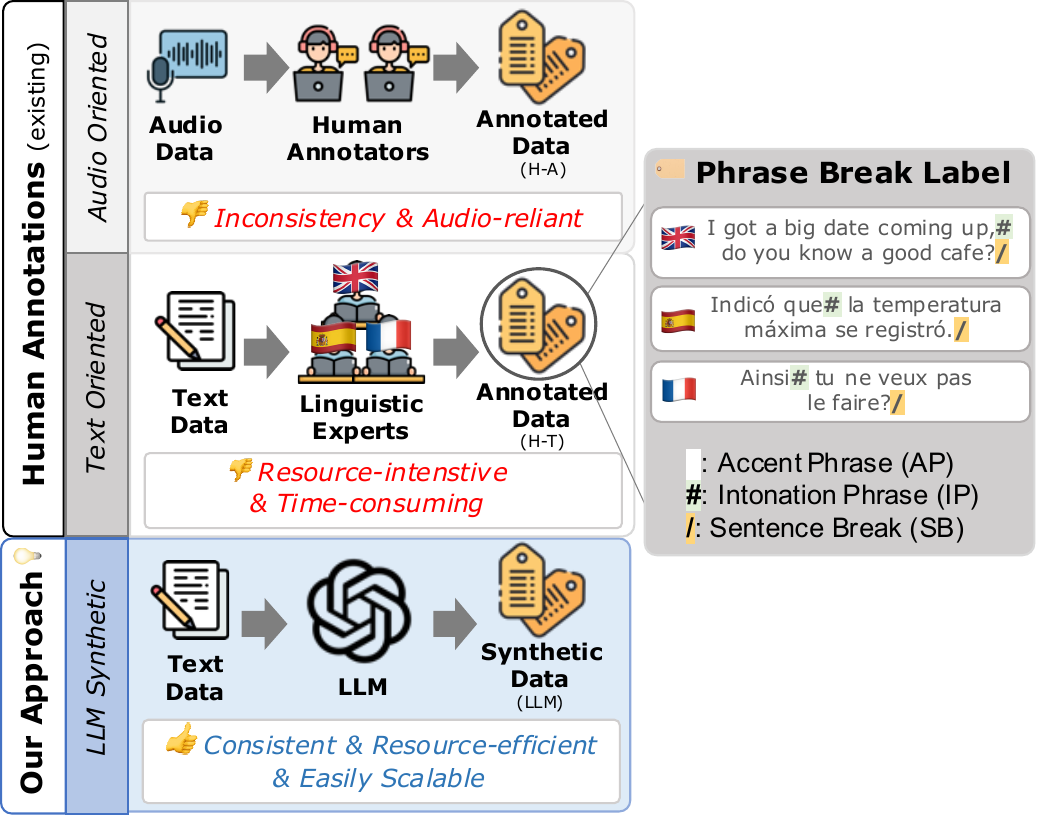}
    \vspace{-0.1in}
    \caption{Overview of traditional human annotations and our LLM-generated synthetic data for phrase break prediction.}
    \label{fig:overview}
  \vspace{-0.26in}
\end{figure}

% 2. Background & Limitation - audio-oriented / text-oriented annotations
We consider two distinct human annotation approaches for marking phrase breaks, as illustrated in Figure~\ref{fig:overview}.
In the first approach, a human annotator identifies phrase breaks directly from the audio by detecting pauses. This approach is known as \textit{audio-oriented} annotations~\cite{Futamata2021,Lee2023}. In the second approach, annotators rely solely on the text, analyzing sentence structure to determine where phrase breaks should occur. These are referred to as \textit{text-oriented} annotations~\cite{taylor1998assigning,klimkov17_interspeech,parlikar2011grammar}.
Although these approaches have been widely used in previous research, each has its own set of challenges. \textit{Audio-oriented} annotations depend on large amounts of high-quality audio data and are prone to inconsistencies caused by variations in recording conditions and phonetic factors~\cite{mao2019nn, meng2010development}.
In contrast, \textit{text-oriented} annotations require extensive linguistic expertise and are considerably more time-consuming to produce.
Furthermore, since both approaches rely on manual human effort, building large-scale, high-quality phrase break datasets requires tremendous temporal and financial resources.
% 2.1. (multilingual TTS 관점에서) 기존 annotations limitation
The heavy reliance on human involvement drives up costs and complexity, a burden further compounded when developing phrase break prediction for multiple languages.

% 3. (multilingual TTS 관점에서) 기존 annotations limitation
% 3. Background - LLM-generated synthetic data in NLP tasks -> Limited in speech domain
Meanwhile, with the remarkable advances in LLMs~\cite{zhouleast,touvron2023llama,achiam2023gpt}, recent studies have focused on exploring their potential and have demonstrated impressive results on a range of downstream tasks.
Among various applications, LLM-based synthetic data generation has emerged as a promising approach to address data scarcity and related challenges, proving effective in generating tailored, high-quality datasets for target tasks.
However, these studies have mainly focused on text-based tasks, like classification~\cite{li-etal-2023-synthetic} and machine translation~\cite{bolding-etal-2023-ask}, while the speech-related domain remains relatively unexplored, despite generally having even more challenging data requirements and constraints.

% 4. Our Approach + 결과 - Section 3,4,5,6
In this paper, we empirically investigate the effectiveness of LLM-based synthetic data generation for phrase break prediction, focusing on addressing the inherent challenges of manual annotation in the speech domain.
We address this question by leveraging an LLM, a text-oriented model, using only a minimal set of examples, significantly less than the thousands typically required in conventional approaches.
In addition, we propose novel strategies based on cross-lingual knowledge transfer to investigate whether this approach can be extended to multiple languages while maintaining minimal resource requirements.
To further assess practical applicability, we train the pretrained multilingual model MiniLM~\cite{wang2020minilm} separately on human and synthetic annotations and analyze the impact on phrase break tasks.

% Contributions - Section 7
Our contributions are summarized as follows:
\begin{itemize}
    \item We present the first empirical study of LLM-generated phrase break annotations, comparing them with two types of traditional human annotations in terms of consistency and alignment with human judgments.
    % \item We provide the first extensive empirical evaluation of LLM-generated phrase break annotations and compare them with traditional human annotations, highlighting differences in annotation consistency and alignment. % our study provides valuable experimental evidence on the potential and limitations of using LLMs to generate synthetic data for phrase break prediction.
    \item Our results demonstrate that LLM-generated annotations exhibit quality comparable to human annotations, while achieving consistent and cost-efficient labeling with significantly fewer samples and greater automation than existing methods.
    % \item By systematically evaluating model performance, we present an efficient alternative to human annotations that significantly reduces both cost and time while maintaining high accuracy.
    \item Our method demonstrates strong generalizability, achieving robust results across multiple languages and validating its applicability in diverse linguistic settings.
\end{itemize}

%% file: contents/2-related.tex
\section{Related Work}
\subsection{Phrase Break Prediction}
Most deep learning-based phrase break models~\cite{pascual2016prosodic,vadapalli2016investigation} demonstrate superior performance over traditional methods~\cite{schmid2004new} by leveraging large-scale human annotations.
These annotations are typically collected through either audio-oriented~\cite{Futamata2021,Lee2023} or text-oriented~\cite{taylor1998assigning,klimkov17_interspeech,parlikar2011grammar} approaches: the former relies on prosodic cues, while the latter uses textual syntax.
Futamata et al.~\cite{Futamata2021} gathered 99,907 audio-oriented annotations for a single language, demonstrating the substantial effort required even in monolingual settings.
While \cite{Lee2023} explored cross-lingual knowledge transfer to reduce annotation costs, their method still required over 60,000 human annotations for the source language.

\subsection{Synthetic Data Generation Using LLMs}
LLMs have excelled in various NLP tasks~\cite{achiam2023gpt, zhouleast, touvron2023llama, li-etal-2023-synthetic, bolding-etal-2023-ask}, particularly in synthetic data generation~\cite{Wang2021, Ding2023}. They effectively address data-related challenges, such as augmenting labeled data for text classification~\cite{li-etal-2023-synthetic} and generating machine translation data~\cite{bolding-etal-2023-ask}.
However, most successes in LLM-based synthetic data generation have been confined to NLP tasks, leaving the speech domain relatively underexplored. 
Unlike text, speech data require prosodic features like rhythm and pitch, making annotation more complex. These challenges become even greater in multilingual settings, where varying prosodic structures and syntactic differences further complicate consistent annotation.

%% file: contents/3-method.tex
\section{Methodology}
\label{sec:3}
In this section, we outline the procedure we followed when leveraging LLM (i.e., the cutting-edge GPT-4o mini~\cite{achiam2023gpt}) to generate phrase break annotations. We assume access to only a minimal set of predefined phrase break labels. We employ a carefully designed system prompt (Figure~\ref{fig:prompt}) that assigns the LLM the persona of a linguistic expert, while a task prompt instructs it to mark phonetic pauses with ``\#'' and sentence boundaries with ``\text{/}'' without altering the original text. To improve accuracy, the LLM’s persona is configured to read sentences thoroughly and speak them aloud before annotation, emulating the human annotation process.

In the first study (Section~\ref{sec:4}), we conduct an exploratory analysis of different annotation types in English as a preliminary step to verify the potential of an LLM-based approach for phrase break generation.
Specifically, we assess the LLM’s ability to follow instructions and validate its annotations (\textbf{LLM}) by comparing them with two distinct human annotation types: \textit{audio-oriented} (\textbf{H-A}) and \textit{text-oriented} (\textbf{H-T}).
Building on this analysis, we evaluate the quality of synthetic annotations in Section~\ref{sec:5}.
To validate LLM-based synthetic annotation beyond English, where resources are typically abundant, we extend our evaluation to French and Spanish, which have fewer resources than English.
We then adapt the prompt for a cross-lingual transfer by switching the LLM's persona from an English expert to a multilingual expert and adding English examples (the most commonly used language in LLM training), into the few-shot examples.
Finally, in Section~\ref{sec:6}, we assess synthetic data's practical value by training a smaller model~\cite{wang2020minilm} on LLM-generated annotations, then assess its performance to determine whether LLM annotations can feasibly replace human annotations.

%% file: contents/4-zerofew.tex
% \section{Evaluation I: Comparative Analysis of Synthetic and Human Annotations} 
\section{Exploratory Analysis: Comparison of Different Annotation Types}
\label{sec:4}

\begin{figure}[t]
    \centering
    \includegraphics[width=0.93\linewidth]{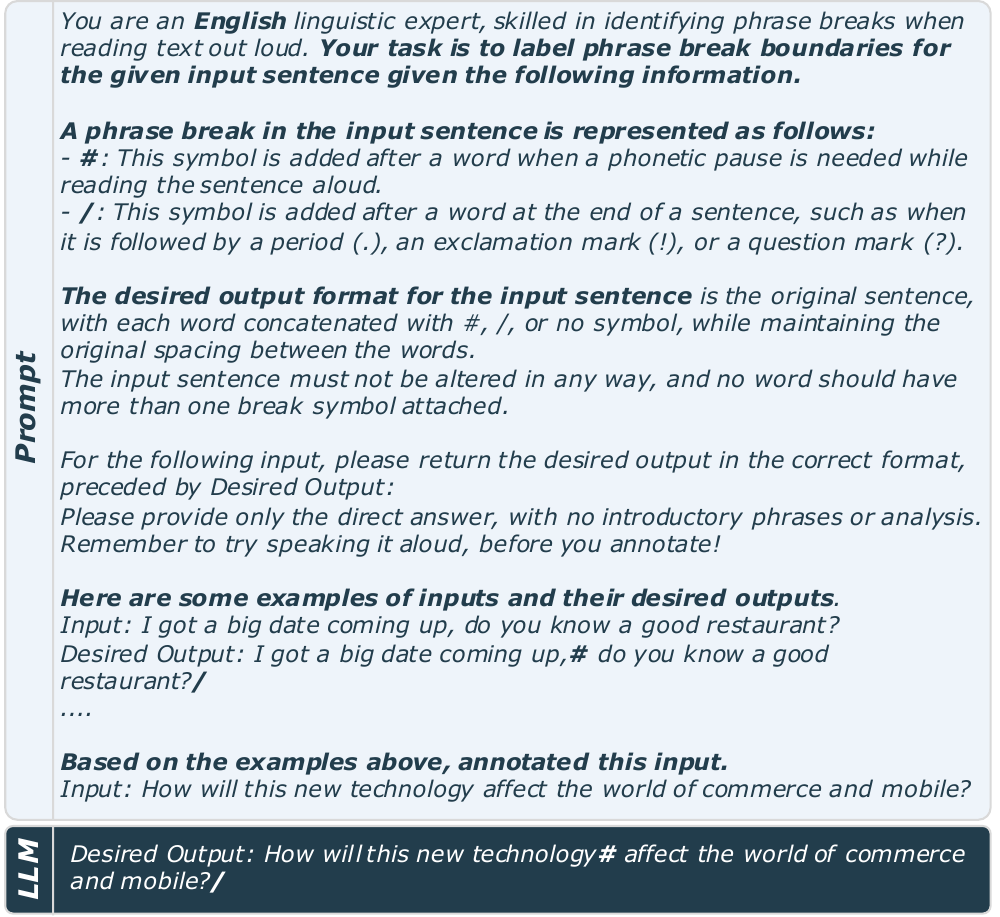}
    \vspace{-0.1in}
    \caption{Prompt for generating phrase break annotation.}
    \label{fig:prompt}
  \vspace{-0.26in}
\end{figure}

\subsection{Experimental Setup}
\textbf{Dataset} We collect 1,000 English utterances from eight native speakers across news, community websites, and spoken corpora to capture varied speech patterns. These are manually annotated in two ways: using recordings (\textbf{H-A}) and by linguistic experts relying on text (\textbf{H-T}). For H-A, an annotator marks breaks based on silences and transitions, following \cite{Futamata2021,Lee2023}.
Each utterance is annotated with three phrase break labels: accent phrase (\textbf{AP}), intonation phrase (\textbf{IP}), and sentence boundary (\textbf{SB}).
% We built a dataset of 1,000 English utterances from diverse domains, spoken by eight native speakers. It includes news, community websites, and spoken corpora to capture varied speech patterns.
% A linguist, independent of the evaluation expert, manually annotated phrase breaks based on silences and transitions in recorded utterances, following \cite{Futamata2021,Lee2023}.

\noindent \textbf{Model \& Prompt}
We use GPT-4o mini\footnote{gpt-4o-mini-2024-07-18} for data generation, given its robust general performance and cost-effectiveness~\cite{wang2025large, schnabel2025multi}. By focusing on a single high-performing LLM, we aim to isolate the effects of different annotation types, enabling a more controlled analysis of their impact on phrase breaks.
We set the temperature to 0.0 and top-p to 1.0, applying the prompt in Figure~\ref{fig:prompt}. The model then generates synthetic annotations in zero-shot (\textbf{ZS}) and few-shot (\textbf{FS}) settings, using expert-annotated utterances from the same domain to maintain consistency.

\noindent \textbf{Metrics} We use pass rate~\cite{iskander2024quality} to measure how well the model follows task instructions in phrase break generation.
Additionally, we employ agreement~\cite{cheng2024autodetect} and Krippendorff’s $\alpha$~\cite{elangovan2024beyond} to measure annotation consistency for the same utterance. Agreement verifies whether all phrase breaks in a sequence match, while Krippendorff's $\alpha$ quantifies inter-annotator reliability.

\begin{table}[t]
    \caption{Annotation comparison by the number of examples ($k$). Pass rate (\%) and average phrasing information per utterance (\%), with standard deviation in parentheses.}
    \label{tab:pass_rate}
    \centering
    \vspace{-0.1in}
    \resizebox{0.95\columnwidth}{!}{%
    \setlength\tabcolsep{5 pt}
    {
    \footnotesize
    \begin{tabular}{ll cccc}
    \toprule
    \multicolumn{2}{c}{\textbf{Annotation Type}} & \textbf{Pass Rate} & \textbf{AP} & \textbf{IP} & \textbf{SB} \\  
    \midrule
    \multirow{2}{*}{\textbf{Human}} & \textbf{H-A} & 100. & 79.0 (11.3) & 9.2 (10.4) & 11.9 (6.1)  \\
                                    & \textbf{H-T} & 100. & 79.3 (8.2)  & 8.8 (7.4)  & 12.0 (6.4)  \\  
    \hline
    \multirow{9}{*}{\textbf{LLM}} & \textit{ZS}        & 98.1  & 18.4 (7.4)  & 72.2 (11.1) & 9.4 (3.9)   \\
                                  & \textit{FS, k=2}   & 99.3  & 86.1 (6.6)  & 4.6 (6.5)   & 9.3 (3.8)   \\
                                  & \textit{FS, k=4}   & 98.5  & 84.8 (6.4)  & 5.9 (6.7)   & 9.3 (3.7)   \\
                                  & \textit{FS, k=8}   & 98.4  & 83.1 (5.9)  & 7.6 (6.4)   & 9.3 (3.7)   \\
                                  & \textit{FS, k=16}  & 98.4  & 82.3 (5.8)  & 8.5 (6.3)   & 9.3 (3.7)   \\
                                  & \textit{FS, k=32}  & 98.5  & 83.2 (5.9)  & 7.5 (6.4)   & 9.2 (3.7)   \\
                                  & \textit{FS, k=64}  & 98.4  & 82.9 (5.9)  & 7.9 (6.3)   & 9.2 (3.7)   \\
                                  & \textit{FS, k=128} & 98.3  & 83.2 (5.8)  & 7.5 (6.3)   & 9.3 (3.8)   \\
                                  & \textit{FS, k=256} & 98.1  & 82.3 (6.4)  & 8.0 (6.6)   & 9.7 (4.1)   \\ 
        \bottomrule
    \end{tabular}}}
    \vspace{-0.1in}
\end{table}

\subsection{Results and Analysis}
\label{sec: 4.2}
\subsubsection{Synthetic Data Generation with LLM}
\label{sec: 4.2.1}
\noindent \textbf{The proposed prompt effectively guides the LLM to follow instructions.}  
To verify whether the LLM correctly followed the prompt instructions, we first examined the pass rate in Table~\ref{tab:pass_rate}. While the distributions of phrase break labels showed similar average values for H-A and H-T, H-T exhibited lower standard deviations in AP and IP, indicating greater consistency.
This variation in H-A could be attributed to differences in recording conditions and speaker characteristics~\cite{kapadia2023multiple}. For synthetic annotations, the pass rate in all settings was comparable to that of human annotations, validating the effectiveness of our prompt.

\noindent \textbf{Few-shot examples improve LLM alignment to human annotations.} 
Although the pass rate remains stable, the distributions of phrasing information reveal notable differences. In the zero-shot setting, synthetic annotations deviate significantly from H-A and H-T, highlighting the difficulty of generating correct labels without annotation examples. As $k$ increases, these values progressively align with human annotations, ultimately converging with H-A and H-T. This demonstrates that incorporating relevant examples helps LLMs better adhere to task-specific annotation methods, thereby improving consistency.

\begin{figure}[t]
    \centering        \includegraphics[width=\linewidth]{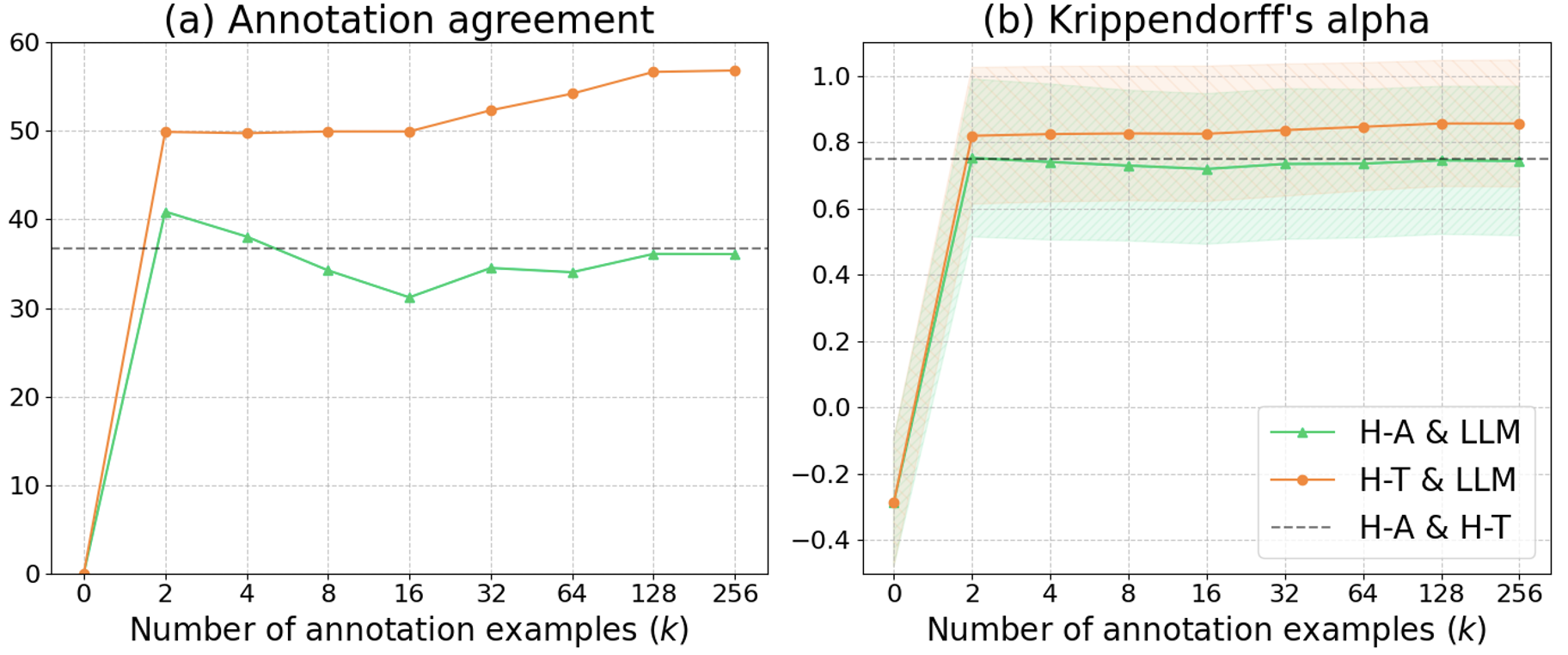}
    \vspace{-0.2in}
    \caption{Agreement (\%) and Krippendorff's $\alpha$ across annotation types by $k$. Shaded regions show standard deviation.}
    \label{fig:en_comparison}
  \vspace{-0.25in}
\end{figure}

\subsubsection{Reliability Comparison of Different Annotation Types}
\label{sec: 4.2.2}
\noindent \textbf{Distinct prosodic cues in H-A and H-T lead to low agreement.}  
We conducted a comparative analysis to examine the correlation between annotation types, as shown in Figure~\ref{fig:en_comparison}. The results show that although H-A and H-T are commonly used as ground truth, their agreement remains relatively low at approximately 36–38\%. This suggests that while both methods capture meaningful phrase break patterns, they emphasize different aspects of prosody—H-A being influenced by natural speech variations~\cite{mao2019nn, meng2010development}, whereas H-T aligns more with syntactic structure~\cite{parlikar2011grammar, klimkov17_interspeech}. Additionally, Krippendorff’s $\alpha$ suggests moderate reliability, reinforcing the notion that H-A and H-T are not entirely interchangeable. Their differences indicate that each provides complementary perspectives.

\noindent \textbf{Can LLMs generate human-Like labels for speech tasks?} 
LLM annotations in the zero-shot setting showed complete misalignment with human annotations (AG = 0, $\alpha\approx-0.28$). Our error analysis confirmed that zero-shot prompting was ineffective, as the LLM misplaced break symbols, overused ``\#'', or misaligned breaks around punctuation.
In contrast, providing even a few examples significantly improved agreement across all annotation types. While H-A \& LLM remained below human agreement for most $k$ values, H-T \& LLM steadily improved, gaining 15–17\% and achieving $\alpha \geq 0.85$, as shown in Figure~\ref{fig:en_comparison}, indicating a highly reliable level of agreement. With sufficient examples, LLM annotations closely align with H-T, suggesting their ability to capture structural prosodic patterns from textual cues, expanding their potential in speech tasks.

\begin{table}[t]
    \caption{Performance of human annotations and LLM annotations in English. Human Score \& F1 when using H-A \& H-T as target labels, with gap from human annotations in parentheses.}
    \label{tab:annotation_result}
    \centering
    \vspace{-0.1in}
    \resizebox{0.95\columnwidth}{!}{%
    \setlength\tabcolsep{5 pt}
    {
    \footnotesize
    \begin{tabular}{llccc}
        \toprule
        \multicolumn{2}{c}{\textbf{Annotation Type}} & \textbf{Human Score} & \textbf{F1 (H-A)} & \textbf{F1 (H-T)} \\  
        \midrule
        \multirow{2}{*}{\textbf{Human}} & \textbf{H-A} & 56.30 & - & 82.14 \\
                                        & \textbf{H-T} & 85.70 & 82.14 & - \\  
        \hline
        \multirow{9}{*}{\textbf{LLM}} & \textit{ZS}        & 0     & 39.31\scriptsize{\textcolor{red}{(-42.8)}} & 39.56\scriptsize{\textcolor{red}{(-42.6)}} \\
                                      & \textit{FS, k=2}   & 71.61 & 79.52\scriptsize{\textcolor{red}{(-2.6)}} & 86.12\scriptsize{\textcolor{ForestGreen}{(+4.0)}} \\
                                      & \textit{FS, k=4}   & 72.50 & 80.13\scriptsize{\textcolor{red}{(-2.0)}} & 87.33\scriptsize{\textcolor{ForestGreen}{(+5.2)}} \\
                                      & \textit{FS, k=8}   & 75.13 & 80.79\scriptsize{\textcolor{red}{(-1.4)}} & 88.31\scriptsize{\textcolor{ForestGreen}{(+6.2)}} \\
                                      & \textit{FS, k=16}  & 76.61 & 80.73\scriptsize{\textcolor{red}{(-1.4)}} & 88.56\scriptsize{\textcolor{ForestGreen}{(+6.4)}} \\
                                      & \textit{FS, k=32}  & 78.61 & 81.17\scriptsize{\textcolor{red}{(-1.0)}} & 88.74\scriptsize{\textcolor{ForestGreen}{(+6.6)}} \\
                                      & \textit{FS, k=64}  & 79.46 & 81.33\scriptsize{\textcolor{red}{(-0.8)}} & 89.55\scriptsize{\textcolor{ForestGreen}{(+7.4)}} \\
                                      & \textit{FS, k=128} & 81.60 & \textbf{81.89\scriptsize{\textcolor{red}{(-0.3)}}} & 90.19\scriptsize{\textcolor{ForestGreen}{(+8.1)}} \\
                                      & \textit{FS, k=256} & \textbf{83.72} & 81.81\scriptsize{\textcolor{red}{(-0.3)}} & \textbf{90.25\scriptsize{\textcolor{ForestGreen}{(+8.1)}}} \\
        \bottomrule
    \end{tabular}}}
    \vspace{-0.22in}
\end{table}

%% file: contents/5-crosslingual.tex
\begin{table*}[t]
\caption{Performance of human and LLM-generated annotations under different cross-lingual few-shot settings.}
    \vspace{-0.1in}
    \centering
    \resizebox{0.9\linewidth}{!}{%
    \footnotesize
\begin{tabular}{llcccccccc}
\toprule
\multicolumn{2}{c}{\multirow{2}{*}{\textbf{Annotation Type}}}                                                                     & \multicolumn{4}{c}{\textbf{\textit{X} = French}}                                                                                                                                           & \multicolumn{4}{c}{\textbf{\textit{X} = Spanish}}                                                                                                                                            \\ \cline{3-10} 
\multicolumn{2}{c}{}                                                                                                              & \textbf{Pass Rate} & \textbf{Human Score} & \textbf{F1 (H-A)}                                                                                    & \textbf{F1 (H-T)}                                                                                    & \textbf{Pass Rate} & \textbf{Human Score} & \textbf{F1 (H-A)}                                                                                    &    \textbf{F1 (H-T)}                                                                                              \\  \midrule
\multirow{2}{*}{\textbf{Human}} & \textbf{H-A}                                                                                            & 100.                                & 67.50                                 & -                                                                                              & 82.91                                                                                          & 100.                                & 53.60                                 & -                                                                                              & 73.56                                                                                          \\  
                                & \textbf{H-T}                                                                                             & 100.                                & 88.20                                 & 82.91                                                                                          & -                                                                                              & 100.                                & 86.70                                 & 73.56                                                                                          & -                                                                                              \\ \hline
\multirow{5}{*}{\textbf{LLM}}   
                                & \textit{FS} (X 16 $\to$ X)                         & 91.1                                & 81.75                                 & 77.19\scriptsize{\textcolor{red}{(-5.7)}}                  & 83.74\scriptsize{\textcolor{ForestGreen}{(+0.8)}}          & 98.8                                & 89.87                                 & \textbf{77.14\scriptsize{\textcolor{ForestGreen}{(+3.6)}}} & \textbf{77.28\scriptsize{\textcolor{ForestGreen}{(+3.7)}}} \\  
                                & \textit{En 4 + X 12} $\to$ X                         & 94.2                                & 78.73                                 & 79.09\scriptsize{\textcolor{red}{(-3.8)}}                  & 85.13\scriptsize{\textcolor{ForestGreen}{(+2.2)}}          & 99.3                                & \textbf{91.14}                        & 74.03\scriptsize{\textcolor{ForestGreen}{(+0.5)}}          & 77.13\scriptsize{\textcolor{ForestGreen}{(+3.6)}}          \\  
                                & \textit{En 8 + X 8} $\to$ X                          & 97.0                                & 84.07                                 & 81.27\scriptsize{\textcolor{red}{(-1.6)}}                  & 86.43\scriptsize{\textcolor{ForestGreen}{(+3.5)}}          & 99.7                                & \textbf{91.14}                        & 72.98\scriptsize{\textcolor{red}{(-0.6)}}                  & 76.96\scriptsize{\textcolor{ForestGreen}{(+3.4)}}          \\  
                                & \textit{En 12 + X 4} $\to$ X                         & 99.7                                & \textbf{84.89}                        & 83.63\scriptsize{\textcolor{ForestGreen}{(+0.7)}}          & \textbf{86.89\scriptsize{\textcolor{ForestGreen}{(+4.0)}}} & 99.8                                & 82.28                                 & 71.28\scriptsize{\textcolor{red}{(-2.3)}}                  & 74.17\scriptsize{\textcolor{ForestGreen}{(+0.6)}}          \\  
                                & \textit{En 16 + X 0} $\to$ X                         & 100.                                & 79.50                                 & \textbf{84.77\scriptsize{\textcolor{ForestGreen}{(+1.9)}}} & 86.02\scriptsize{\textcolor{ForestGreen}{(+3.1)}}          & 100.                                & 70.90                                 & 71.55\scriptsize{\textcolor{red}{(-2.0)}}                  & 73.48\scriptsize{\textcolor{red}{(-0.1)}}                  \\  
\bottomrule
\end{tabular}}
    \label{tab:cross_lingual}
    \vspace{-0.2in}
\end{table*}

% \section{Evaluation I: Comparison Across Different Languages}
\section{Evaluation I: Annotation Quality}
\label{sec:5}
\subsection{Experimental Setup}
\textbf{Dataset} 
Following the same procedure used for English data in Section~\ref{sec:4}, we collect 1,000 utterances each for French and Spanish, along with two corresponding human annotations. The same phrase break labels and domains apply across all settings.
% Following the same settings as the English data in Section~\ref{sec:4}, we collect 1,000 utterances each for French and Spanish. These datasets also serve as the source for few-shot examples, annotated by French and Spanish linguistic experts.

\noindent\textbf{Model \& Prompt} 
Most of the model configurations from Section~\ref{sec:4} are maintained, with a few key modifications. Considering real-world constraints where obtaining language-specific few-shot examples is challenging, we fix \(k=16\). We exclude the zero-shot setting due to its significant deviation from human annotations, and include the cross-lingual setting which transfers knowledge from the source language.
In the cross-lingual setting, we use annotation examples from both English (\textit{En}) as the source language and target language \textit{X}. We set the total number of annotation examples from \textit{En} and \textit{X} to 16.
% Specifically, we allocate \( k_s \in \{4, 8, 12, 16\} \) for \textit{En} and \( k_x \in \{12, 8, 4, 0\} \) for \textit{X}, ensuring a total of \( k_s + k_x = 16 \).

% The model and hyperparameters follow Section~\ref{sec:4}, with an extended prompt for cross-lingual few-shot settings. Given the time and cost constraints associated with limited  availability of experts in multiple languages, LLM-based multilingual data generation offers a viable solution. As shown in Section~\ref{sec:4}, performance at $k=16$ was comparable to $k=256$ against H-T with only 16 few-shot examples. Thus, we adopt a realistic low-resource scenario, setting $k=16$, distributing examples inversely between English ($k_{eng}=4, 8, 12, 16$) and target languages ($k_{trg}=12, 8, 4, 0$).

\noindent\textbf{Metrics}
We introduce a human score where a linguistic expert conducts a binary evaluation (acceptable/unacceptable) on each text-annotation pair. To avoid bias, these evaluators differ from those who created the few-shot examples and human annotations. Following \cite{Lee2023}, we adopt the macro-averaged F1 score.
% We use the human score, where evaluators assess text-annotation pairs via binary Pass/Fail evaluation. To prevent potential bias, these evaluators are distinct from those who created the few-shot examples and human annotations. We adopt the macro-averaged F1-score, ensuring a balanced assessment~\cite{Lee2023}.
 % We conduct qualitative and quantitative assessments to evaluate our method on English and multilingual synthetic datasets.

\subsection{Results and Analysis}
\subsubsection{Human vs. Synthetic Annotations}
\noindent \textbf{H-T serves as a more consistent ground truth.} To assess annotation quality from a human perspective, we evaluated human score in Table~\ref{tab:annotation_result}. The results show a gap of approximately 30\% between H-A and H-T, which does not indicate superiority but rather reflects H-T’s consistency with human evaluation, while H-A captures speech variability. As noted in Section~\ref{sec: 4.2.2}, H-A is influenced by speaker traits like intonation and speaking rate, leading to inconsistencies. Thus, H-T’s higher score reflects its structured consistency, indicating that H-T suits structured tasks while H-A captures natural speech patterns. 

\noindent \textbf{How can LLMs generate human-like labels?}
When compared with H-A, the F1 consistently increased as more shots were provided, approaching 82.14. Additionally, when evaluated against H-T, the LLM exceeded H-A’s F1 of 82.14 at \(k=2\), indicating stronger alignment with text-oriented annotations. However, both human score and F1-score plateaued after a certain \(k\), with F1-score gains reaching saturation and human score slightly declining at higher \(k\).  
These results suggest that few-shot examples help LLMs infer prosodic cues for phrase break, extending their applicability to speech tasks. The LLM-generated annotations aligned with H-T while maintaining reasonable agreement with H-A, balancing text and speech prosody. Thus, selecting and optimizing few-shot examples is essential for maximizing performance, underscoring the role of prompt design in capturing speech-driven variations.

\subsubsection{Annotations in Other Languages}
\label{sec: 5.2.2}
\noindent\textbf{Linguistic differences affect cross-lingual data generation.}  
Incorporating English data improved pass rate and human score, with language-specific effects. In French, human score peaked at \textit{En12 + X4}, demonstrating effective knowledge transfer (Table~\ref{tab:cross_lingual}). Even at \textit{En16 + X0}, human score dropped to 79.50 but still exceeded H-A's human score.  
For Spanish, a balanced ratio was crucial, with the highest human score at \textit{En8 + X8}. With excessive English data, the human score dropped to 70.90 at \textit{En16 + X0}. 
These differences stem from linguistic factors. Adding English data enhances annotation quality since French and English share structural similarities~\cite{liu2014comparative, vander2018prosodic}. In contrast, Spanish, a syllable-timed language, differs from stress-timed English, leading to distinct phrase break patterns~\cite{colantoni2014learning}. While English data benefits Spanish, excessive reliance disrupts prosodic alignment, lowering human scores. This pattern is consistent with how LLMs model multiple languages, capturing shared structures while maintaining language-specific distinctions~\cite{zhang2024same}. In cross-lingual scenarios, using target language data with English data better enhances the quality of annotations than relying solely on target language data.

%% file: contents/6-xlb.tex
% \section{Exploratory Analysis: Impact of Annotations on Model Performance}
\section{Evaluation II: Impact of Annotations\\on Model Performance}
\label{sec:6}

\subsection{Experimental Setup}
\textbf{Dataset}
We use the same human and LLM annotations for each language (Sections~\ref{sec:4} and~\ref{sec:5}).
%  from both few-shot and cross-lingual settings / as described
Given real-world constraints, we set \textit{k} to 16, and to 256 for resource-rich English.
For cross-lingual few-shot (\textbf{XL-FS}) settings, we use two configurations: balanced 
\(En8 + X8\) and English-only \(En16 + X0\).
% (\textit{En}8 + \textit{X}8) and English-only (\textit{En}16 + \textit{X}0).
% Considering real-world constraints, we primarily set k to 16, while also including k=256 for English where abundant resources are typically available.
% For cross-lingual settings, we utilize annotations from two configurations: a balanced setting with 8 examples each from English and target language, and a resource-constrained setting using only English examples.
% We use the same human-annotated datasets, H-T and H-A, as described in Sections~\ref{sec:4} and~\ref{sec:5}. For LLM annotations, we apply few-shot prompting in English, whereas for French and Spanish, we use few-shot and cross-lingual few-shot settings on 1,000 utterances. 
We collect utterances for generation input from CommonVoice~\cite{ardila-etal-2020-common}, a widely used open-domain speech dataset with broad coverage.
% , which enhances generalizability
% Evaluation is conducted on CommonVoice~\cite{ardila-etal-2020-common}, an open-domain speech dataset chosen due to its broad coverage, which ensures generalizability.

\noindent\textbf{Training} 
We divide the annotation data into train, validation, and test sets with a ratio of 85:7.5:7.5.
For phrase break prediction, we employ MiniLM\footnote{https://huggingface.co/microsoft/Multilingual-MiniLM-L12-H384}, a distilled multilingual language model capable of learning from each language's annotations.
The model achieving the highest macro-F1 score on the test set is chosen for evaluation.
Two human annotations, H-A~\cite{Futamata2021,Lee2023} and H-T~\cite{taylor1998assigning,klimkov17_interspeech,parlikar2011grammar}, serve as baselines for comparison with LLM-generated annotations.
% The train, validation, and test data are split into an 85:7.5:7.5 ratio. We select MiniLM\footnote{https://huggingface.co/microsoft/Multilingual-MiniLM-L12-H384}, a distilled multilingual language model capable of learning from each language’s annotations.
% We choose MiniLM\footnote{https://huggingface.co/microsoft/Multilingual-MiniLM-L12-H384}, which outperformed DistilmBERT~\cite{sanh2019distilbert, devlin-etal-2019-bert} despite its smaller size.
% The model with the highest macro-F1 score on test set during training is selected as the final model. Baseline models are trained using H-A and H-T annotations for comparison with LLM-based phrase break generations. Notably, the baselines use the same settings as~\cite{Futamata2021,Lee2023}.

\subsection{Results and Analysis}
Table~\ref{tab: baseline} presents macro-F1 scores on the test set and human scores on CommonVoice.
The model trained on LLM annotations achieved the highest F1 in English, suggesting that lower subjectivity and greater consistency enhance performance.
% Table~\ref{tab: baseline} presents the results, indicating the few-shot model in English achieved the highest F1, surpassing human annotations. 
% This suggests that the dataset's lower subjectivity and higher consistency improve phrase break generations.
Similarly, the French model achieved the highest F1, confirming the effectiveness of few-shot prompting. Notably, in the \textit{En16 + Fr0} setting, LLM annotation exceeded both H-A and H-T in terms of F1. For Spanish, the \(k=16\) model achieved an F1 comparable to human annotations. Likewise, the model trained on synthetic data outperformed the human-annotated model when evaluated on the CommonVoice. In general, LLM annotations steadily outperformed human-labeled data in F1, validating that few-shot prompting enables effective phrase break generation. Furthermore, the results indicate that English examples offer meaningful support for phrase break generation, reinforcing the findings from Section~\ref{sec: 5.2.2}. Notably, this approach proved generalizable, achieving high performance in French and Spanish, confirming its cross-linguistic applicability.

\begin{table}[t]
    \caption{Evaluation of the impact of trained annotations.}
    % \caption{Performance of baseline models trained on H-A\&H-T and the proposed methods.}
    \label{tab: baseline}
    \centering
    \vspace{-0.1in}
    \resizebox{0.95\columnwidth}{!}{%
    \setlength\tabcolsep{5 pt}
    {
    \footnotesize
    \begin{tabular}{l l c c}
        \toprule
        \textbf{Language} & \textbf{Annotation Type} & \textbf{F1} & \textbf{CommonVoice} \\
        \midrule
        \multirow{4}{*}{English} 
        & \textbf{H-A} & 82.55 & 58.5 \\
        & \textbf{H-T} & 89.89 & 72.2 \\
        & \textbf{LLM} \textit{(FS, k=16)} & \textbf{94.90} & 64.6 \\
        & \textbf{LLM} \textit{(FS, k=256)} & 94.51 & \textbf{75.7} \\
        \midrule
        \multirow{5}{*}{French} 
        & \textbf{H-A} & 90.32 & 76.2 \\
        & \textbf{H-T} & 91.93 & 82.8 \\
        & \textbf{LLM} \textit{(FS, k=16)} & \textbf{98.59} & 69.0 \\
        & \textbf{LLM} \textit{(XL-FS, En 8 + Fr 8)} & 96.09 & 78.4 \\
        & \textbf{LLM} \textit{(XL-FS, En 16 + Fr 0)} & 94.06 & \textbf{84.5} \\
        \midrule
        \multirow{5}{*}{Spanish} 
        &\textbf{H-A} & 89.25 & 55.1 \\
        & \textbf{H-T} & 89.57 & 56.1 \\
        & \textbf{LLM} \textit{(FS, k=16)} & \textbf{89.84} & \textbf{69.4} \\
        & \textbf{LLM} \textit{(XL-FS, En 8 + Es 8)} & 88.33 & \textbf{69.4} \\
        & \textbf{LLM} \textit{(XL-FS, En 16 + Es 0)} & 89.04 & 57.1 \\
        \bottomrule
    \end{tabular}}}
    \vspace{-0.25in}
\end{table}

%% file: contents/7-conclusion.tex
\section{Conclusion}
In this paper, we present a comprehensive study on synthetic data generation for phrase break prediction, leveraging LLMs to address fundamental challenges in the speech domain arising from phonetic factors.
Our results demonstrate that synthetic phrase break annotations effectively overcome the limitations of conventional human annotations, which are hampered by extensive resource requirements and inconsistencies due to human involvement.
We also find that models trained on synthetic data can perform comparable or even superior to those trained with human annotations derived from audio and text sources.
Our work highlights the potential for cost-efficient data generation with minimal examples, offering a promising solution even for resource-limited languages.